# A PID-Controlled Tensor Wheel Decomposition Model for Dynamic Link Prediction

***Qu Wang*[1]*, and Yan Xia*[2*]**

[1]*College of Computer and Information Science, Southwest University, Chongqing, China*
[2]*School of Electronic Information and Communication Engineering, Chongqing Aerospace Polytechnic Chongqing, China*
**Email address of corresponding author: xiayan_99@163.com*

## Abstract

Link prediction in dynamic networks remains a fundamental challenge in network science, requiring the inference of potential interactions and their evolving strengths through spatiotemporal pattern analysis. Traditional static network methods have inherent limitations in capturing temporal dependencies and weight dynamics, while tensor-based methods offer a promising paradigm by encoding dynamic networks into high-order tensors to explicitly model multidimensional interactions across nodes and time. Among them, tensor wheel decomposition (TWD) stands out for its innovative topological structure, which decomposes high-order tensors into cyclic factors and core tensors to maintain structural integrity. To improve the prediction accuracy, this study introduces a PID-controlled tensor wheel decomposition (PTWD) model, which mainly adopts the following two ideas: 1) exploiting the representation power of TWD to capture the latent features of dynamic network topology and weight evolution, and 2) integrating the proportional-integral-derivative (PID) control principle into the optimization process to obtain a stable model parameter learning scheme. The performance on four real datasets verifies that the proposed PTWD model has more accurate link prediction capabilities compared to other models.

## 1 Introduction

Link prediction in dynamic networks is a key issue in network science and complex system analysis [1-8]. Its core goal is to infer potential or missing links and their weights through the dynamic evolution of network structure and attributes. With the rapid growth of real network data such as social networks, transportation systems, and biological protein interactions, the network not only presents dynamic changes in the connection state between nodes, but also contains continuous fluctuations in the strength of connections [9-12]. However, due to the limitations of information collection technology or data transmission loss, the actual observed network often lacks some link information, which has a significant impact on downstream tasks such as network structure analysis, community discovery, and influence propagation prediction [13, 14]. Traditional link prediction methods are mostly based on the assumption of static or unweighted networks, and it is difficult to effectively capture the spatiotemporal correlation of the dynamic evolution of weights. Especially in the case of co-evolution of network topology and weights, it is urgent to develop new prediction models that integrate temporal features, weight heterogeneity, and network high-order structures [15-21].

In recent years, methods based on tensor network have gradually become an important direction for solving this problem [22-26]. A 3rd-order tensor can naturally represent the temporal evolution of the connection status and weights between nodes in a dynamic network through two node dimensions and one time dimension [27-30]. Then, through tensor network technology, the model can mine potential low-dimensional patterns from it and achieve accurate prediction of missing links and their weights. As one of the most classic tensor network forms, CP decomposition has received extensive attention for link prediction. For example, Luo *et al*. [31] design a novel linear bias of CP decomposition in the form of rank-one tensor to simulate the data fluctuation. Wu *et al*. [32] introduce a fine-grained regularization scheme to enhance the generalization ability of the CP decomposition model. Luo *et al*. [33] adopt the ADMM framework to model the non-negative constraints under CP decomposition, which improved the optimization efficiency of the model. The above models are all based on CP tensor network, which have the problem of insufficient representation ability due to the limitation of latent feature space. Some other types of tensor network models have been proposed to improve the accuracy of link prediction. Tang *et al*. [34] propose a nonnegative tucker decomposition model for predicting missing links in QoS data. Wang *et al*. [35] use tensor ring decomposition to design a link prediction model with a larger representation space. Although the above models based on tucker decomposition and tensor ring decomposition have achieved good performance, the improved accuracy is limited due to its structural limitations [36-46].

As shown in Fig.1, Tensor wheel decomposition (TWD) represents a high-order tensor by combining a core tensor and a ring topology tensor, and has a more powerful representation capability [47]. In order to obtain accurate link prediction results, this paper proposes a PID-controlled tensor wheel decomposition (PTWD) model. The core idea is to model the dynamic network as a third-order tensor and capture the high-order spatiotemporal correlation of the interaction between nodes through TWD. In order to further optimize the stability and convergence efficiency of model training, the PID control mechanism is introduced to dynamically adjust the instantaneous error, historical error accumulation and future trend prediction during the parameter update process, effectively suppressing the oscillation problem of stochastic gradient descent. Experiments show that PTWD significantly improves the link prediction accuracy. This paper's main contributions are:

- A link prediction model is established using TWD to obtain more accurate prediction results.
- The PID control is integrated into the TWD's parameter learning scheme to obtain a stable training process.

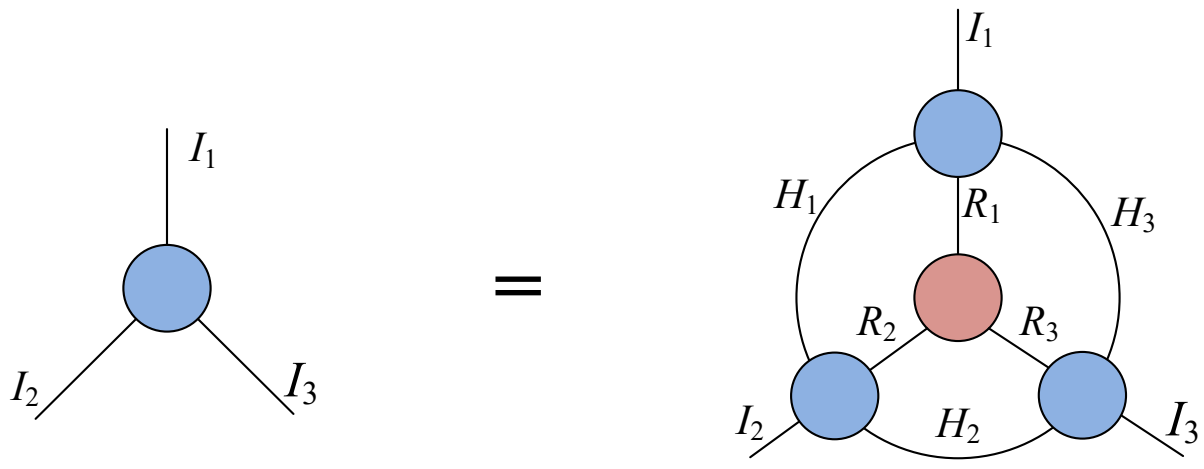

Fig. 1 Tensor wheel decomposition.

## 2. Preliminaries

***Definition 1 (A 3rd-order tensor modeling* dynamic networks*).*** The third-order tensor modeling of dynamic weighted networks is a structured representation method for time-series interaction data. Its core idea is to integrate the node, time and weight information in the network into a unified third-order tensor. Specifically, the model maps the interaction intensity of each node pair in the network at different time segments to the corresponding elements of a three-dimensional array, where the uncollected data is marked as a special missing state. Under this framework, the evolution of node behavior patterns over time can be mined through tensor decomposition technology, thereby realizing the prediction of missing links and weights.

***Definition 2 (Tensor Wheel Decomposition, TWD).*** TWD is an innovative decomposition method based on tensor networks. Its core principle is to decouple high-order tensor structures into two key components: a set of fourth-order cyclic tensors and a core tensor, and to achieve collaborative associations between components through wheel-like topology. The model captures the implicit cyclic dependency patterns in multidimensional data through cyclic tensors, while using core tensors to model global interaction features, and finally reconstructs the low-rank approximation $\hat{\mathbf{Y}}$ of the original tensor through tensor reduction operations $TW(*)$:

$$\hat{\mathbf{X}} = TW[\mathbf{G};\mathbf{A},\mathbf{B},\mathbf{C}], \tag{1}$$

where $\mathbf{G}\in\mathbb{R}^{H1\times H2\times H3}$ is a 3rd-order core tensor, $\mathbf{A}\in\mathbb{R}^{R3\times|I|\times R1\times H1}$, $\mathbf{B}\in\mathbb{R}^{R1\times|J|\times R2\times H2}$, $\mathbf{C}\in\mathbb{R}^{R2\times|K|\times R3\times H3}$ are three 4th-order factor tensors. Then, according to the principle of tensor contraction, we can get the calculation method of a single element in tensor $\hat{\mathbf{Y}}$:

$$\hat{x}_{ijk} = \sum_{r_1=1}^{R_1}\sum_{r_2=1}^{R_2}\sum_{r_3=1}^{R_3}\sum_{h_1=1}^{H_1}\sum_{h_2=1}^{H_2}\sum_{h_3=1}^{H_3} g_{h_1h_2h_3}a_{r_3ir_1h_1}b_{r_1jr_2h_2}c_{r_2kr_3h_3}. \tag{2}$$

## 2. Methodology

When tensor decomposition is used for reconstruction, the Euclidean distance is usually used to measure the difference between the original tensor $\mathbf{Y}$ and the low-rank approximation tensor $\hat{\mathbf{Y}}$, and the optimization objective function is constructed based on this. Therefore, the following objective function is obtained by combining the single-element calculation formula (2) of TWD:

$$\begin{aligned}\varepsilon &= \left\|\mathbf{X}-\hat{\mathbf{X}}\right\|_F^2 = \sum_{i=1}^{|I|}\sum_{j=1}^{|J|}\sum_{k=1}^{|K|}\left(x_{ijk}-\hat{x}_{ijk}\right)^2 \\ &= \sum_{i=1}^{|I|}\sum_{j=1}^{|J|}\sum_{k=1}^{|K|}\left(x_{ijk}-\sum_{r_1=1}^{R_1}\sum_{r_2=1}^{R_2}\sum_{r_3=1}^{R_3}\sum_{h_1=1}^{H_1}\sum_{h_2=1}^{H_2}\sum_{h_3=1}^{H_3} g_{h_1h_2h_3}a_{r_3ir_1h_1}b_{r_1jr_2h_2}c_{r_2kr_3h_3}\right)^2.\end{aligned} \tag{3}$$

Note that (3) requires traversing each element in the tensor $\mathbf{Y}$, which is very time-consuming for a large-scale dynamic network. Combining the idea of data density orientation [48-52], focusing on the known entries in the tensor $\mathbf{Y}$ during the modeling process can greatly reduce the computational complexity. Therefore, a data density-oriented TWD-based objective function is:

$$\begin{aligned}\varepsilon &= \sum_{x_{ijk}\in\Lambda}\left(x_{ijk}-\hat{x}_{ijk}\right)^2\\ &= \sum_{x_{ijk}\in\Lambda}\left(x_{ijk}-\sum_{r_1=1}^{R_1}\sum_{r_2=1}^{R_2}\sum_{r_3=1}^{R_3}\sum_{h_1=1}^{H_1}\sum_{h_2=1}^{H_2}\sum_{h_3=1}^{H_3} g_{h_1h_2h_3}a_{r_3ir_1h_1}b_{r_1jr_2h_2}c_{r_2kr_3h_3}\right)^2.\end{aligned}\tag{4}$$

In the training process of tensor decomposition models, in order to prevent the model from overfitting the noise or accidental patterns in the observed data, regularization terms are often introduced to constrain the parameter space [53-65]. Specifically, $L_2$ regularization suppresses the abnormal growth of parameter values by constraining the sum of squares of elements in the factor tensor, thereby reducing model complexity and enhancing generalization ability[66-69]. At this time, the loss function with regularization terms can be defined as:

$$\begin{aligned}\varepsilon &= \sum_{x_{ijk}\in\Lambda}\left(\left(x_{ijk}-\hat{x}_{ijk}\right)^2+\lambda\left(\|\mathbf{G}\|_2^2+\|\mathbf{A}\|_2^2+\|\mathbf{B}\|_2^2+\|\mathbf{C}\|_2^2\right)\right).\\ &= \sum_{x_{ijk}\in\Lambda}\left(x_{ijk}-\sum_{r_1=1}^{R_1}\sum_{r_2=1}^{R_2}\sum_{r_3=1}^{R_3}\sum_{h_1=1}^{H_1}\sum_{h_2=1}^{H_2}\sum_{h_3=1}^{H_3} g_{h_1h_2h_3}a_{r_3ir_1h_1}b_{r_1jr_2h_2}c_{r_2kr_3h_3}\right)^2\\ &+\lambda\sum_{x_{ijk}\in\Lambda}\left(\sum_{r_1=1}^{R_1}\sum_{r_2=1}^{R_2}\sum_{r_3=1}^{R_3} g_{h_1h_2h_3}^2+\sum_{r_1=1}^{R_1}\sum_{r_3=1}^{R_3}\sum_{h_1=1}^{H_1} a_{r_3ir_1h_1}^2\right)\\ &+\lambda\sum_{x_{ijk}\in\Lambda}\left(\sum_{r_1=1}^{R_1}\sum_{r_2=1}^{R_2}\sum_{h_2=1}^{H_2} b_{r_1jr_2h_2}^2+\sum_{r_2=1}^{R_2}\sum_{r_3=1}^{R_3}\sum_{h_3=1}^{H_3} c_{r_2kr_3h_3}^2\right),\end{aligned}\tag{5}$$

where λ is the regularization coefficient, which is used to adjust the balance between data fitting accuracy and model complexity. For the optimization problem of objective function (5), the stochastic gradient descent (SGD) algorithm can be used to solve it [70-76]. In the iterative process, the instantaneous error of sample $y_{ijk}$ in the nth training round is defined as $e_{ijk}^n=y_{ijk}^n-\hat{y}_{ijk}^n$, and its mathematical form is isomorphic to the proportional adjustment term in the classic PID controller. At this time, the sample information only affects the parameter update through the single-order feedback of the error term $e_{ijk}$, which is essentially equivalent to a simplified PID control structure with proportional gain coefficient $C_P$=1, integral coefficient $C_I$=0 and differential coefficient $C_D$=0. However, in order to further improve the convergence speed and stability of model training, this study draws on the idea of error dynamic compensation in control theory, introduces the integral term and differential term of the complete PID controller into the learning process, and constructs a composite error correction mechanism. The improved instantaneous error expression is adjusted to:

$$\tilde{e}_{ijk}^n = C_P e_{ijk}^n + C_I\sum_{f=1}^{n} e_{ijk}^f + C_D\left(e_{ijk}^n - e_{ijk}^{n-1}\right).\tag{6}$$

Among them, the three coefficients are used to adjust the weight distribution of the current error, historical error integral and error change rate [77-81]. This error modeling strategy that integrates the dynamic characteristics of the time series can effectively suppress the parameter oscillation phenomenon caused by local gradient noise in traditional SGD [82-85]. Therefore, the following update formula is obtained based on the PID-guided SGD algorithm:

$$\begin{aligned} g_{h_1h_2h_3} &\leftarrow g_{h_1h_2h_3}+\eta\left(\tilde{e}_{ijk}\sum_{r_1=1}^{R_1}\sum_{r_2=1}^{R_2}\sum_{r_3=1}^{R_3} a_{r_3ir_1h_1}b_{r_1jr_2h_2}c_{r_2kr_3h_3}-\lambda g_{h_1h_2h_3}\right),\\ a_{r_3ir_1h_1} &\leftarrow a_{r_3ir_1h_1}+\eta\left(\tilde{e}_{ijk}\sum_{r_2=1}^{R_2}\sum_{h_2=1}^{H_2}\sum_{h_3=1}^{H_3} g_{h_1h_2h_3}b_{r_1jr_2h_2}c_{r_2kr_3h_3}-\lambda a_{r_3ir_1h_1}\right),\\ b_{r_1jr_2h_2} &\leftarrow b_{r_1jr_2h_2}+\eta\left(\tilde{e}_{ijk}\sum_{r_3=1}^{R_3}\sum_{h_1=1}^{H_1}\sum_{h_3=1}^{H_3} g_{h_1h_2h_3}a_{r_3ir_1h_1}c_{r_2kr_3h_3}-\lambda b_{r_1jr_2h_2}\right),\\ c_{r_2kr_3h_3} &\leftarrow c_{r_2kr_3h_3}+\eta\left(\tilde{e}_{ijk}\sum_{r_1=1}^{R_1}\sum_{h_1=1}^{H_1}\sum_{h_2=1}^{H_2} g_{h_1h_2h_3}a_{r_3ir_1h_1}b_{r_1jr_2h_2}-\lambda c_{r_2kr_3h_3}\right),\end{aligned}\tag{7}$$

where $\eta$ represents the learning rate. So far, we have obtained a PTWD model with high accuracy and stable convergence process.

## 3 Experiment and Result Analysis

### *3.1 Experimental Preparation*

***Datasets.*** This paper uses four real dynamic network datasets, the details of which are shown in Table 1. D1-D2 are collected from a metropolitan area network in a certain place, and the interaction intensity represents the size of the data packet transmitted between devices. D3-D4 are from a cryptocurrency trading website, and the interaction intensity represents the transaction amount between the two parties. In order to stabilize the training process, we uniformly normalize the interaction intensity [86-91], i.e., $\log(x_{ijk}+1)$.

Table 1 Dataset explanation

| Datasets | Nodes | Time slot | Edge | Density |
|---|---|---|---|---|
| D1 | 240,500 | 765 | 658,079 | $1.49\times10^{-8}$ |
| D2 | 128,805 | 580 | 208,894 | $2.17\times10^{-8}$ |
| D3 | 198,863 | 124 | 563,420 | $1.15\times10^{-7}$ |
| D4 | 98,022 | 120 | 206,980 | $1.80\times10^{-7}$ |

***Experimental setup.*** To ensure the fairness of the experiment, all models were written in JAVA and run on a computer with i7-13700CPU and 32GRAM. In addition, we divided the data set into training set, validation set and test set in a ratio of 1:2:7, and in order to avoid chance, we divided the data ten times to take the average of the results.

***Evaluation indicators.*** In order to evaluate the performance of the model, the widely used RMSE and MAE are used as evaluation indicators [92-98]. The smaller the RMSE or MAE, the higher the prediction accuracy.

$$\text{RMSE} = \sqrt{\frac{\sum_{y_{ijk}\in\Omega}\left(y_{ijk}-\widehat{y}_{ijk}\right)^2}{|\Omega|}}, \text{MAE} = \frac{\sum_{y_{ijk}\in\Omega}\left|y_{ijk}-\widehat{y}_{ijk}\right|}{|\Omega|}.$$

**Baselines.** This paper uses five models for comparison, including M1 (tucker decomposition), M2 (tensor ring decomposition), M3[99], M4[100], M5[101], and M6(PTWD). All models are fine-tuned using the parameters used in their articles to ensure the best accuracy.

### *3.2 Results and Discussion*

For fair comparison, the latent feature dimensions of all models are uniformly set to 5. Then, according to previous experience [3], the PID parameters of PTWD were adjusted to $C_P$=1, $C_I$=0, and $C_D$=0.001. The experimental results are shown in Tables 2-3, and we can get the following findings:

Table 2 RMSE of all tested models on D1-D4

| Datasets | D1 | D2 | D3 | D4 |
|---|---|---|---|---|
| M1 | 0.3068 | 0.2791 | 0.6669 | 0.7418 |
| M2 | 0.2971 | 0.2718 | 0.6323 | 0.6862 |
| M3 | 0.3112 | 0.2857 | 0.6875 | 0.7546 |
| M4 | 0.3180 | 0.2875 | 0.7071 | 0.7607 |
| M5 | 0.3073 | 0.2851 | 0.6650 | 0.7430 |
| M6 | **0.2947** | **0.2703** | **0.6356** | **0.6919** |

Table 3 MAE of all tested models on D1-D4

| Datasets | D1 | D2 | D3 | D4 |
|---|---|---|---|---|
| M1 | 0.2052 | 0.2141 | 0.4362 | 0.5427 |
| M2 | 0.2013 | 0.2202 | 0.4248 | 0.5132 |
| M3 | 0.2005 | 0.2115 | 0.4483 | 0.5473 |
| M4 | 0.2056 | 0.2143 | 0.4527 | 0.5506 |
| M5 | 0.2072 | 0.2208 | 0.4472 | 0.5549 |
| **M6** | **0.1971** | **0.2103** | **0.4216** | **0.5168** |

PTWD, i.e., M6 has higher link prediction accuracy on the four dynamic network datasets. Specifically, as shown in Table 2, the RMSE achieved by M6 on D1 is 0.2947, which is 3.94% lower than 0.3068 achieved by M1, 0.81% lower than 0.2971 achieved by M2, 5.30% lower than 0.3112 achieved by M3, 7.33% lower than 0.3180 achieved by M4, and 4.10% lower than 0.3073 achieved by M5. The RMSE on D2-D4 is similar. In addition, as shown in Table 3, the MAE of M6 on D1 is 0.1971, which is 3.95% lower than 0.2052 obtained by M1, 2.09% lower than 0.2013 obtained by M2, 1.70% lower than 0.2005 obtained by M3, 4.13% lower than 0.2056 obtained by M4, 4.87% lower than 0.2072 obtained by M5. Similarly, similar results can be found on the MAE of D2-D4. Among them, M1 is the model of Tucker decomposition, M2 is the model of tensor ring decomposition, and the M6 we proposed uses tensor wheel decomposition, which combines the advantages of Tucker and tensor ring, uses a core tensor to establish the connection of the ring topology tensor, and can capture the connection between multiple dimensions. M6 can naturally obtain higher link prediction accuracy compared with M1 and M2. Similarly, M3 is a non-negative model based on Tucker decomposition, and naturally achieves worse results than M6. M4 and M5 are both models based on CP decomposition, which are limited by the small latent feature space, resulting in limited representation learning ability of the model.

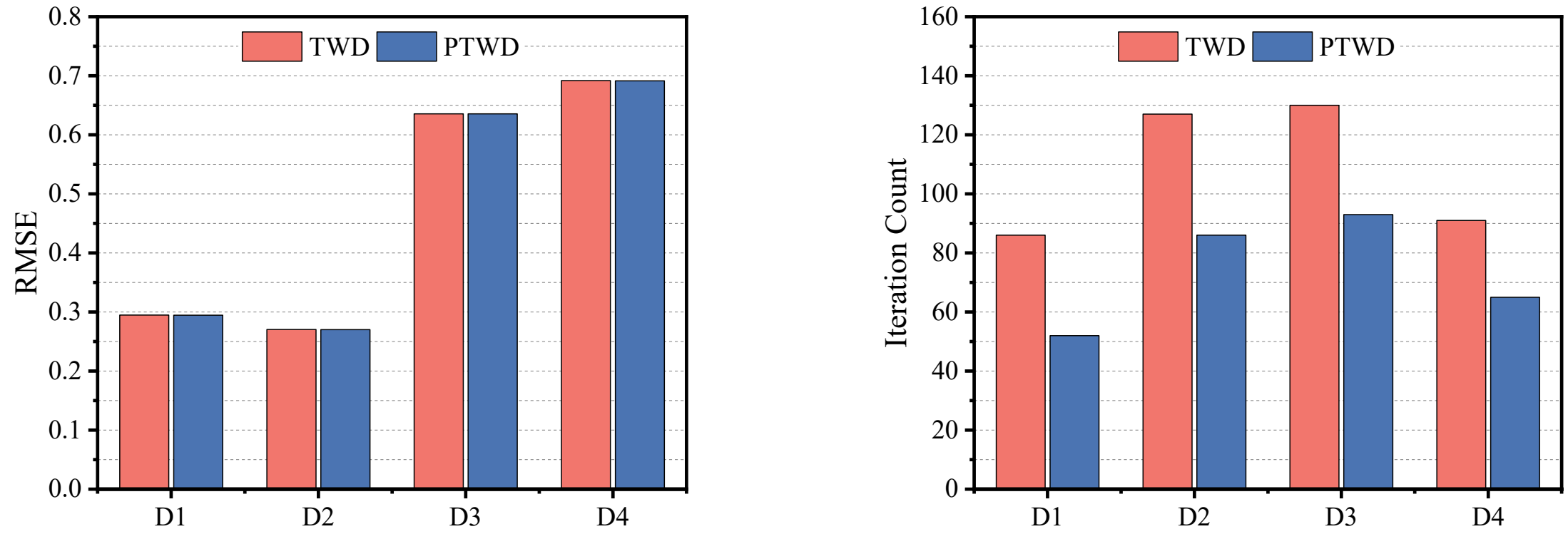

Fig. 2 Results of ablation experiments.

### *3.3 Ablation Study*

In this section, an ablation experiment is performed to verify the role of PID. Specifically, the PID is removed to obtain the TWD model, and then the TWD and PTWD accuracy and number of iterations are observed on the RMSE of D1-D4. The results are recorded in Figure 2. We can find that the accuracy of the proposed model is very little affected by PID, with only a slight improvement. However, there is a huge improvement in the number of iterations, and the model with PID guidance (PTWD) only needs fewer iterations to reach convergence. This also verifies the role of the PID-guided parameter learning plan we designed, which can make the model converge more stably and faster.

## 4 Conclusion

This study proposes a PID-controlled tensor wheel decomposition model (PTWD), which decouples the third-order dynamic network tensor into a combination of ring factors and core tensors through the wheel topology of TWD, effectively capturing the high-order spatiotemporal coupling characteristics of node interactions. Then, PID control theory is further introduced to significantly improve the stability and convergence efficiency of model training by dynamically adjusting the proportional, integral and differential errors in the parameter update process. Experiments on four real data sets show that the prediction accuracy of PTWD in sparse observation scenarios is significantly improved compared with existing methods. In the future, we consider applying the PTWD model to some real application scenarios to test its practical application value [102-108].

## 5 Acknowledgements

This work is supported by the Science and Technology research project of Chongqing Aerospace Polytechnic 2024-XJKJ-04, and the Science and Technology Research Program of Chongqing Municipal Education Commission under grant KJQN202403017.